\documentclass{llncs}
\usepackage{makeidx}  

\usepackage{graphicx}
\usepackage{csvsimple}
\usepackage[export]{adjustbox}
\usepackage{multirow}
\usepackage[table,xcdraw]{xcolor}
\usepackage{float}
\begin{document}
\frontmatter          
\pagestyle{headings}  
\addtocmark{Hamiltonian Mechanics} 

\title{A Paradigm Shift: Detecting Human Rights Violations Through Web Images}
\titlerunning{A Paradigm Shift}  
%

\author{Grigorios Kalliatakis \and Shoaib Ehsan 
\and Klaus D. McDonald-Maier}
\authorrunning{Grigorios Kalliatakis et al.} 
%
%
\institute{School of Computer Science and Electronic Engineering, University of Essex, UK\\
\email{\{gkallia, sehsan, kdm\}@essex.ac.uk}
}

\maketitle              

\begin{abstract}
The growing presence of devices carrying digital cameras, such as mobile phones and tablets, combined with ever improving internet networks have enabled ordinary citizens, victims of human rights abuse, and participants in armed conflicts, protests, and disaster situations to capture and share via social media networks images and videos of specific events. This paper discusses the potential of images in human rights context including the opportunities and challenges they present. This study demonstrates that real-world images have the capacity to contribute complementary data to operational human rights monitoring efforts when combined with novel computer vision approaches. The analysis is concluded by arguing that if images are to be used effectively to detect and identify human rights violations by rights advocates, greater attention to gathering task-specific visual concepts from large-scale web images is required.
\end{abstract}
\section{Introduction}
Digital technologies provide new openings for human rights advocates, including non-governmental organizations (NGOs) such as Amnesty International and Human Rights Watch, to identify, expose, verify and document human rights violations \cite{9}. These emerging technologies are likely to complement conventional interview-based fact-finding approaches in order to effectively verify abuses. In this paper, we use the term and concept of `human rights violations' in keeping up with the ways it has being adopted by human rights advocates, that is, to refer to actions executed by state or non-state actors that breach any part of those rights that protect individuals and groups from behaviours that intervene with fundamental freedoms and human dignity \cite{1}.

Digital images hold immense potential for monitoring emerging human rights violations and emphasising on specific events that might otherwise elude attention. Organizations concerned with human rights advocacy are increasingly using digital images as a tool for improving the exposure of human rights and international humanitarian law violations that may otherwise be impossible. However, these organizations continue to depend mainly on qualified imagery analysts to manually identify human rights violations events from the extensive amount of data. Because of the high expense involved concerning these analysts \cite{2}, very few organizations are able to conduct monitoring campaigns over large-scale data.

Until now, all human rights event detection efforts are implemented by utilising tools such as remote sensing platforms and social media. The aim of this paper is to discuss the value of digital images within the context of detecting human rights violations, in order to analyse major opportunities and difficulties for their growing and prolonged use in supporting human rights. Our research lens is focused on establishing a new approach for human rights advocates to monitor different forms of human rights violations without requiring the use of trained imagery analysts.

The remainder of the paper is organised as follows. Section 2 examines prior works related to human rights monitoring tasks throughout remote sensing and social media. Section 3 examines the value of real-world images for detecting human rights violations. Also in the same section the difficulties of collecting real-world images from the web are thoroughly discussed. Finally, conclusions and future directions are given in section 4.

\section{Background}
Human rights monitoring is an extensive term outlining the collection, verification, and use of information to address human rights problems. The need of regularly detecting phenomena associated with some violations has led organizations to conduct various human rights monitoring campaigns. Up to now, the process of detecting international law violations engages solely two main mechanisms, \textit{remote sensing} and \textit{social media} alongside eyewitness reports. This section will briefly review those two research directions emphasizing on their shortcomings. 

\subsection{Remote Sensing}
Remote sensing is considered a mechanism capable of improving the process of detecting human rights violations and is nowadays performed by platforms such as helicopters and unmanned aerial vehicles where people are at risk of suffering from such offences. Organizations that monitor human rights violations have increasingly followed a typical sequence, as described in \cite{2}, of remote sensing. First the violation which they are seeking to detect needs to be determined. After that, a remotely sensed phenomenon associated with that particular violation is chosen and finally a suitable sensor that will allow the detection of this phenomenon is established. Satellites can be utilised to detect phenomena such as extensive changes to villages \cite{3}, identifying mass graves and large-scale destruction of civilian properties. However, most phenomena currently being exploited in human rights monitoring campaigns, demand a high-resolution sensor. Another drawback can be found when dealing with specific atrocities, such as genocide, where a systematic and a time-series analysis is required for the evidence. The potential for satellite images to be used in human rights contexts has not been fully realised yet, mainly because of the required cost and limited resources.

\subsection{Social Media}
Another mechanism suited for information-gathering tasks by human rights organizations is social media. Such organizations, including NGOs, are currently analysing data from social media, such as Twitter and Facebook, to a greater extent in order to recognise, verify and document different human rights abuses. Social media are capable of circulating information earlier and more comfortably compared to traditional media. Furthermore, social media manage to provide evidence of different types of human rights violations that can avoid being censored by restrictive governments. However, manually identifying human rights violations events from such an extensive amount of data poses serious technical challenges for human rights NGOs. In order to address this challenge, \cite{4} proposed an approach for automated, large-scale analysis of human rights related events. They model social media (such as Twitter) data as a heterogeneous network with multiple different nodes, relationships and features by employing a Non-Parametric Heterogeneous Graph Scan (NPHGS), in order to achieve human rights events identification and characterization from the emerging patterns.

\section{Images for Detecting Human Rights Violations}
Images have grown into omnipresent on the Internet, something that has inspired the already highly motivated process of analysing their semantic content for various applications. From this perspective, digital images have growing usage among the human rights sector. Such organizations are progressively attempting to exploit the capacity of digital images as a tool for identifying, exposing and verifying different forms of human rights and international humanitarian law abuses. As discussed in Section 2.1 there are several limitations for human rights-focused organisations when utilising remote sensing such as: 1) high cost for acquiring high-resolution images; 2) restricted access to valuable resources; 3) limited efficacy of satellite images which leads only to large-scale human rights violations detection; 4) the obligation for qualified imagery analysts and 5) specific sensor characteristics as prerequisites. 
Such difficulties can be overcome by utilising a combination of public available digital images and a novel context-based vision system capable of detecting human rights violations. The most promising aspect of real-world digital images compared to satellite images is that they can be effortlessly captured by using everyday devices such as mobile phones. Consequently, the cost for acquiring such images is practically negligible while their number can rapidly expand without the need of commercial providers.

\subsection{Digital Images as Legal Evidence}
Certainly, the efficiency of publicly available real-world images varies according to purpose and the nature of the images collected. Gathering such evidence can serve not only for advocacy purposes but potentially for evidence in legal cases as well. Evidence is anything that can provide information about an incident being investigated as stated in \cite{5} and may come from many different resources like physical, testimony, documents and imagery. Our future research will focus on real-world images in order to exploit the high-standard characteristics of legal evidence to human rights monitoring process. Typically there are three main questions set by every human rights monitoring mechanism: \emph{what}, \emph{who} and \emph{how}. Examples of such crimes include the following: 1) Military forces torturing a civilian; 2) Bulldozers  unlawfully destroying homes; 3) Children in military camps being trained for warfare and 4) Police placing a suspect in an illegal chokehold. Also documenting the who and how, often referred to as \textit{linkage evidence}, can provide important clues to investigators. For example 1) the face of a soldier as he repeatedly beats a civilian; 2) the face of the bulldozer operator as he destroys homes and 3) the badge number and name plate of an officer who is illegally holding a suspect.  

\begin{figure}[!t]
\centering
\includegraphics[height=7.5cm,width=10.50cm,keepaspectratio,center]{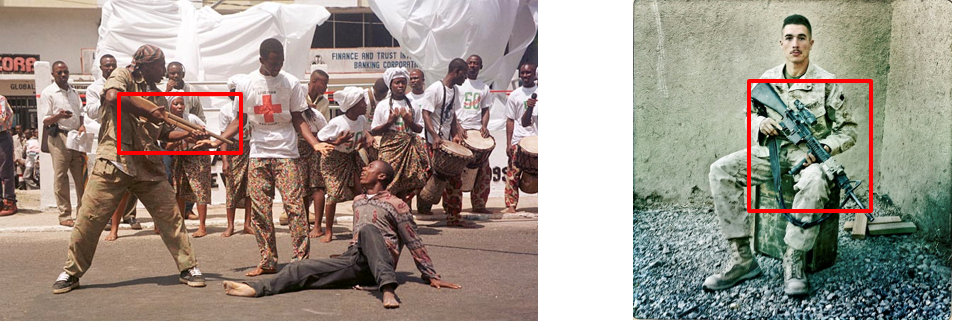}
\caption{Object recognition results for two images. Both images contain a gun but it is evident that only the left image can be utilised for the task of recognising human rights violations.}
\label{fig_sim}
\end{figure}

Object recognition in computer vision is the task of finding and identifying objects in images or videos sequences \cite{10,11}. Humans perform this task effortlessly and instantaneously; however, it remains an exceptionally difficult activity to be outlined from the machines. The object recognition problem is closely engaged to the segmentation problem, while it can also be defined as a labelling problem based on model of known objects. Figure 1 illustrates an example of object recognition for two images.

Image classification is the task of categorizing images into classes according to their depicted contents \cite{12,13}. Large-scale image classification has received serious interest from the computer vision and machine learning communities \cite{6,7}. Figure 2 demonstrates a conventional example of classifying two images. 

If we conduct a further more detailed inspection of the Figures 1-2, the results concerning the two typical aforementioned vision-based problems can easily be deduced for both set of images. However if we were about to answer the following question: \textit{is there a human right being violated in these images?} will both set of images draw the same conclusion? Apparently the answer to this question is totally different for each image. Regarding Figure 1, while both images contain a gun, only in the left image there is a human right being violated which is `right to life' and `freedom from torture and inhuman treatment'. Figure 2 demonstrates classification results with both images being labelled as `camp'. However, the actual situation is quite contrasting with the right image displaying a summer camp, while on the other hand the left image depicts a refugee camp.

\begin{figure}[!t]
\centering
\includegraphics[height=7.5cm,width=10.50cm,keepaspectratio,center]{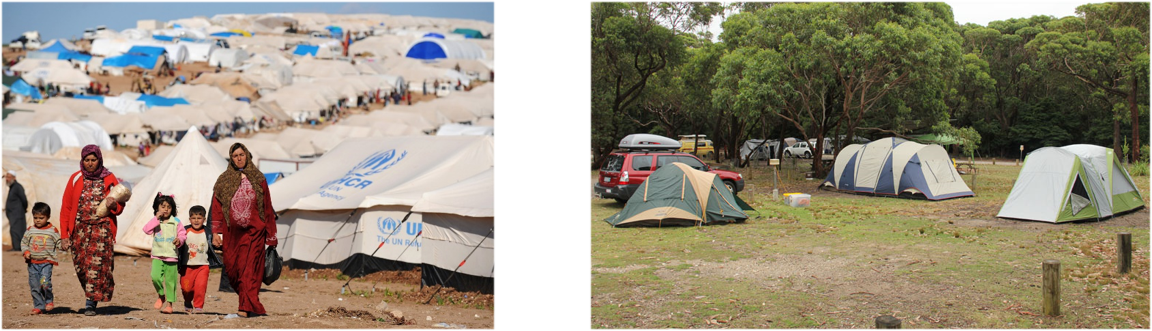}
\caption{Example of image classification results for two images. They are both carrying the label of 'camp', but it is clear that the image on the right cannot be considered for the task of recognising human rights violations.}
\label{fig_sim}
\end{figure}
For these reasons, detecting and identifying human rights violations through digital images cannot be considered as a trivial problem, since individual tasks need to be combined in order for the system to be able to accurately answer the question whether there is a human right being violated in a specific image or not. 
Previously this role was played by qualified human rights imagery analysts. Recently, representation learning methods and essentially convolutional neural networks (CNNs) \cite{8} have enjoyed great success in context-based image understanding.
Consequently, obtaining effective high-level representations from large-scale web images has become increasingly important, while a key question arises in the context of human rights understanding: how will we gather this structured visual knowledge? This section describes the difficulties encountered when collecting those high-level representations for the task of detecting human rights violations.

\begin{figure*}[]
\centering
\includegraphics[width=1\textwidth]{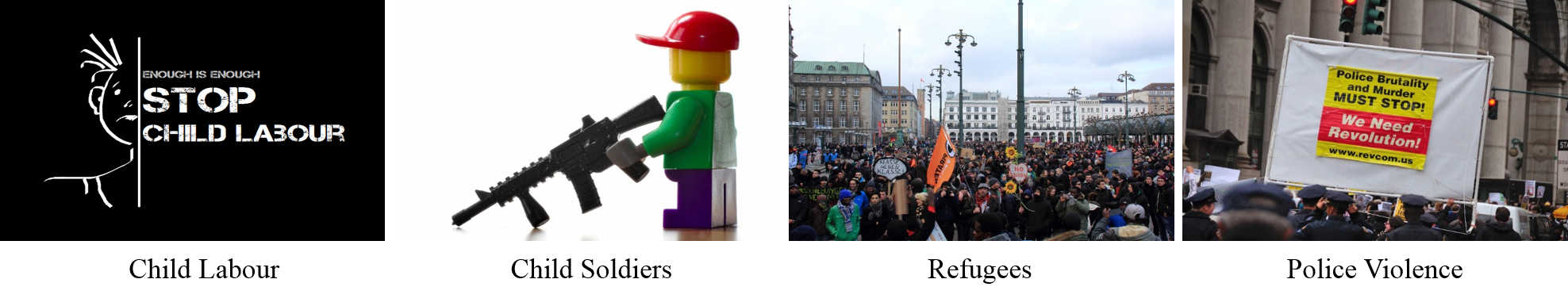}
\caption{Examples of irrelevant images returned from Flickr photo-sharing website alongside the given query term.}
\label{HRUN_examples}
\end{figure*}

\subsection{Preliminary Research and Discussion}
To our knowledge, there is no reference point in standardized dataset of images and annotations regarding human rights violations. The basic component of our intended research is to construct a well-sampled image database in the domain of human rights understanding before exploring the capacity of CNNs at detecting and identifying human rights violation through digital images. 
The first serious issue arise when Flickr photo-sharing website was chosen in order to collect images related to specific keywords assembled by experts on the field of human rights. Figure 3 displays an example of irrelevant results returned for some given queries. 
There have been cases where the given keyword was `armed conflict' and most of the returned images had to do with military parades. Another similar example was when the given keyword was 'refugees' and the returned results included protesting campaigns for refugees, something that may be consider close to the keyword but can not serve our purpose.
Google and Bing search engines were chosen in order to overcome the aforementioned difficulties. However, quickly it became apparent that there was a huge number of irrelevant results returned for the given queries especially when compared with typical inquires such as `car', `aeroplane', `bike', `dog' etc. Table 1 illustrates the percentage of relevant images from the total number of images retrieved for each query concerning four human rights violations categories compared to everyday image queries.
Images were downloaded for each class using a python interface to the Google and Bing application programming interfaces (APIs), with the maximum number of images permitted by their respective API for each query term. While the process may be repeated several times in order to get additional images, for the purpose of this comparison a single attempt is presented. This is due to the limited amount of images returned for the rest of the attempts that were conducted with regard to human rights violations. It is evident from Table 1 that in order to obtain real-world images, which can be regarded as appropriate for the creation of a dataset in the human rights violations detection context, greater effort and vast time are required compared to other conventional datasets.

\begin{table*}[]
\centering
\caption{Statistics of the image collection procedure from search engines. First column displays the query terms. Next, the number of retrieved images alongside the number of applicable images is presented. The last column indicates the quantitative relation between those two sets of images.}
\label{my-label}
\resizebox{\textwidth}{!}{
\begin{tabular}{|ccccccc|}
\hline
\rowcolor[HTML]{C0C0C0} 
\cellcolor[HTML]{C0C0C0}                                         & \multicolumn{2}{c}{\cellcolor[HTML]{C0C0C0}\textbf{Retrieved  Images}} & \multicolumn{2}{c}{\cellcolor[HTML]{C0C0C0}\textbf{Relevant  Images}} & \multicolumn{2}{c|}{\cellcolor[HTML]{C0C0C0}\textbf{Ratio}} \\
\rowcolor[HTML]{C0C0C0} 
\multirow{-2}{*}{\cellcolor[HTML]{C0C0C0}\textbf{Query Term}} & \textbf{Google}                    & \textbf{Bing}                    & \textbf{Google}                    & \textbf{Bing}                   & \textbf{Google}                       & \textbf{Bing}                      \\ \hline
child labour                                       & 99                                 & 137                              & 18                                 & 5                               & 18\%                                  & 3.64\%                             \\
child soldiers                                                   & 176                                & 159                              & 31                                 & 13                              & 17.61\%                               & 8.17\%                             \\
police violence               & 149                                & 232                              & 10                                 & 16                              & 6.71\%                                & 6.89\%                             \\
refugees                                                         & 111                                & 140                              & 10                                 & 39                              & 9\%                                   & 27.85\%                            \\ \hline
aeroplane                                                        & 170                                & 137                              & 150                                & 135                             & 88.23\%                               & 98.54\%                            \\
car                                                              & 145                                & 128                              & 123                                & 124                             & 84.82\%                               & 96.87\%                            \\
dog                                                              & 105                                & 132                              & 101                                & 129                             & 96.19\%                               & 97.72\%                            \\ \hline
\end{tabular}}
\end{table*}

\section{Conclusions and Future Work}
Recognising   and   identifying   human   rights   violations   through   digital   images   is   a   unique   and challenging problem in the field of computer vision. The overall goal of this paper is to investigate by what means semantically meaningful content must be extracted from digital images in order to highlight those that can be considered for human rights monitoring purposes.
The results of this preliminary research indicate that obtaining  suitable  high-level  representations  in  the  context  of  human rights violations, is a demanding task. Additionally, what is clear from investigating the capacity of real-world images in the context of human rights understanding is that human rights advocates can  be  supported  in  the  difficult  tasks  of  verifying  and  tracking human rights violations,  eye-witness testimony, prosecuting and other statements on human rights violations in greater extent than previous high cost approaches. Inspired by the great success of representation learning methods and principally CNNs, our future research will focus on tackling the untrodden problem of detecting and recognising human rights violations throughout real-world images by exploiting these novel techniques.

\section*{Acknowledgements}
This work was supported by the Economic and Social Research Council [grant number ES/M010236/1]. 

%
%

\clearpage
\end{document}